\documentclass{article}

\usepackage{microtype}
\usepackage{graphicx}
\usepackage{subcaption}
\usepackage{booktabs} % for professional tables

\usepackage{hyperref}

\usepackage{placeins}

\usepackage[preprint]{icml2026}

\usepackage{amsmath}
\usepackage{amssymb}
\usepackage{mathtools}
\usepackage{amsthm}
\usepackage{float}

\usepackage[capitalize,noabbrev]{cleveref}
\usepackage{bbm}

\definecolor{mask}{RGB}{78,149,217}
\definecolor{occlusion}{RGB}{233,133,50}
\definecolor{hmr}{RGB}{78,167,46}

\theoremstyle{plain}

\theoremstyle{definition}

\theoremstyle{remark}

\usepackage[textsize=tiny]{todonotes}

\usepackage{xspace}
\newcommand{\method}{SAM-Body4D\xspace}

\begin{document}

\twocolumn[
  \icmltitle{SAM-Body4D: Training-Free 4D Human Body Mesh Recovery from Videos}

  % It is OKAY to include author information, even for blind submissions: the
  % style file will automatically remove it for you unless you've provided
  % the [accepted] option to the icml2026 package.

  % List of affiliations: The first argument should be a (short) identifier you
  % will use later to specify author affiliations Academic affiliations
  % should list Department, University, City, Region, Country Industry
  % affiliations should list Company, City, Region, Country

  % You can specify symbols, otherwise they are numbered in order. Ideally, you
  % should not use this facility. Affiliations will be numbered in order of
  % appearance and this is the preferred way.
  \icmlsetsymbol{equal}{*}

  \begin{icmlauthorlist}
    \icmlauthor{Mingqi Gao}{sheffield}
    \icmlauthor{Yunqi Miao}{warwick}
    \icmlauthor{Jungong Han}{tsinghua}
    % \icmlauthor{Firstname4 Lastname4}{sch}
    % \icmlauthor{Firstname5 Lastname5}{yyy}
    % \icmlauthor{Firstname6 Lastname6}{sch,yyy,comp}
    % \icmlauthor{Firstname7 Lastname7}{comp}
    % %\icmlauthor{}{sch}
    % \icmlauthor{Firstname8 Lastname8}{sch}
    % \icmlauthor{Firstname8 Lastname8}{yyy,comp}
    %\icmlauthor{}{sch}
    %\icmlauthor{}{sch}
  \end{icmlauthorlist}

  \icmlaffiliation{tsinghua}{Department of Automation, Tsinghua University, Beijing, China}
  \icmlaffiliation{warwick}{University of Warwick, Coventry, UK }
  \icmlaffiliation{sheffield}{University of Sheffield, Sheffield, UK }

  \icmlcorrespondingauthor{Jungong Han}{jungonghan77@gmail.com}
  % \icmlcorrespondingauthor{Firstname2 Lastname2}{first2.last2@www.uk}

  % You may provide any keywords that you find helpful for describing your
  % paper; these are used to populate the "keywords" metadata in the PDF but
  % will not be shown in the document
  \icmlkeywords{Machine Learning, ICML}

  \vskip 0.3in
]

% this must go after the closing bracket ] following \twocolumn[ ...

% This command actually creates the footnote in the first column listing the
% affiliations and the copyright notice. The command takes one argument, which
% is text to display at the start of the footnote. The \icmlEqualContribution
% command is standard text for equal contribution. Remove it (just {}) if you
% do not need this facility.

% Use ONE of the following lines. DO NOT remove the command.
% If you have no special notice, KEEP empty braces:
\printAffiliationsAndNotice{}  % no special notice (required even if empty)
% Or, if applicable, use the standard equal contribution text:
% \printAffiliationsAndNotice{\icmlEqualContribution}

\begin{abstract}
Human Mesh Recovery (HMR) aims to reconstruct 3D human pose and shape from
2D observations and is fundamental to human-centric understanding in
real-world scenarios. While recent image-based HMR methods such as SAM~3D~Body achieve
strong robustness on in-the-wild images, they rely on per-frame inference when
applied to videos, leading to temporal inconsistency and degraded performance
under occlusions. We address these issues without extra training by leveraging the inherent human continuity in videos.
We propose \method, a training-free framework for temporally consistent and
occlusion-robust HMR from videos. We first generate identity-consistent
masklets using a promptable video segmentation model, then refine them with an
Occlusion-Aware module to recover missing regions. The refined masklets guide
SAM~3D~Body to produce consistent full-body mesh trajectories, while a
padding-based parallel strategy enables efficient multi-human inference.
Experimental results demonstrate that \method achieves improved
temporal stability and robustness in challenging in-the-wild videos, without
any retraining. Our code and demo are available at:
\url{https://github.com/gaomingqi/sam-body4d}.
\end{abstract}    
\section{Introduction}
\label{sec:intro}

\begin{figure*}[t]
  \centering
  \includegraphics[width=1\linewidth]{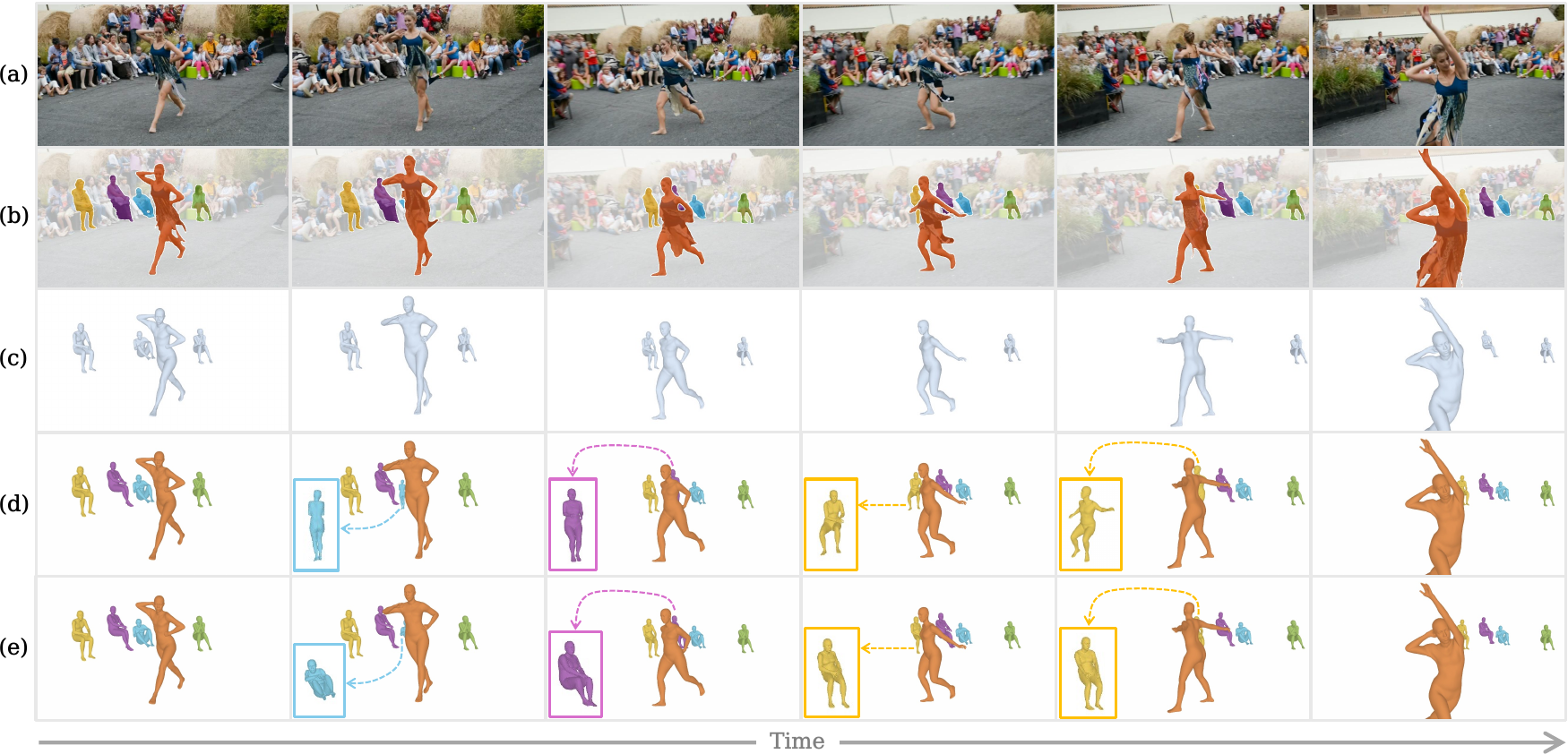}
  \caption{Illustration of temporally consistent Human Mesh Recovery (HMR) from
videos. (a) Input video frames. (b) Identity-consistent human masks, where each
person is highlighted with a unique and consistent colour across frames. (c)
Vanilla image-to-video HMR baseline using SAM~3D~Body with automatic human
detection and per-frame inference. Note that only the meshes corresponding to
the masks in (b) are visualised here; if a mesh does not appear in a certain
frame, it indicates that the corresponding person is not detected in that
frame. (d) Our spatial-temporal consistent HMR, where the temporal continuity
in masklets is directly propagated into the 4D human meshes. (e) Our full
\method with occlusion-aware refinement. Across the 2nd–5th columns,
\method recovers plausible and temporally stable reconstructions under
occlusion. As these humans
are heavily occluded, their complete meshes are visualised in the bottom-left
corner for clearer observation.}
% (e.g., the blue person in the 2nd column, the purple person in the 3rd column, and the yellow person in the 4th and 5th columns)
  \label{fig:illustration}
\end{figure*}

% Human Mesh Recovery 定义和意义

Human Mesh Recovery (HMR) aims to reconstruct 3D human pose and shape from
2D visual observations, offering explicit representations of human geometry.
These representations are valuable for numerous human-centric applications,
including human–robot interaction, behaviour analysis, sports performance
understanding, immersive VR/AR environments, and embodied AI. 

% Motivation, SAM 3D Body 只在 image 上，video / in-the-wild videos (promptable & occlusions) 做图片的很多，但是，image to video 会有很多 complex 的 challenge，我们做 video。

Recent advances in image-based HMR have significantly improved generalization in in-the-wild conditions. 
Most notably, SAM~3D~Body~\cite{sam3dbody} achieves
strong robustness through a scalable data engine, a separate optimization strategy for body and hands, and a decoupled skeleton/shape representation via
Momentum Human Rig (MHR~\cite{mhr}), resulting in more reliable performance than SMPL/SMPL-X-based methods on diverse and complex images. 
However, when extended to videos, most image-based HMR methods operate in a frame-by-frame manner, relying heavily on independent human detection results for each input image. 
As per-frame detections lack temporal continuity, the reconstructed human meshes often fluctuate and fail to remain stable in video scenarios (Fig.~\ref{fig:illustration}(c)).
This becomes particularly problematic in in-the-wild settings where dynamic camera motion, background clutter, and frequent occlusions often cause mixed identities and tracking breakdowns.

% 以前方法的两个问题：
% 从video本身出发，不是从结果出发
% optimization bases
Although video-based HMR methods attempt to ensure temporal continuity by modeling temporal information~\cite{vibe} or incorporating tracking mechanisms,~\cite{tram}, they are fundamentally optimization based, which demands large annotated video datasets and carefully crafted objectives.
Such reliance restricts their scalability and weakens the robustness to diverse and unpredictable in-the-wild human motions and scene dynamics.
% In this work, we introduce a fully training-free framework for human mesh recovery from in-the-wild videos, enabling stable 4D human reconstruction without any video-specific training.

% 利用 video continuty 提升 4D 重构连续性，Training-free 的基础

% 怎么保证 Training-Free 的, 我们利用了 SAM-3 的时间连续性，能提供连续 masK，是 fine-grained 参考

% Although video-based HMR methods attempt to exploit temporal information, most of their modeling focuses on feature or pose space, leaving the strong human-level continuity already present in the 2D pixels under-utilised. Videos naturally preserve coherent spatio-temporal structures for each human at the pixel level. We explicitly capture this fine-grained continuity through identity-consistent masklets and propagate it into the reconstructed 4D meshes, enabling temporal stability to be grounded in direct visual evidence and supporting a fully training-free paradigm for video HMR.

% Our framework（image 到 video 扩展）（平滑过渡）（数量固定，有连续性，用 SAM-3 的理由）（相机视角的移动 & 遮挡）
% Training-Free

% 介绍 SAM 3D Body
Building upon this insight, we propose \method, a training-free framework
for temporally consistent HMR from videos. Given an input video, \method
generates identity-preserving mesh trajectories for the target humans. We first
track and segment the target pixels using a promptable video segmentation
model, producing identity-consistent masklets that carry temporal continuity.
These masklets then serve as prompts to guide SAM~3D~Body, transferring the
inherent temporal coherence of videos to the reconstructed 4D human meshes.
Furthermore, an Occlusion-Aware Refiner is introduced to recover missing or
corrupted regions caused by occlusions, preventing hallucinated predictions.
Additionally, a padding-based parallel strategy enables efficient multi-person
and multi-frame inference without modifying pre-trained models. An overview of
our framework is shown in Fig.~\ref{fig:illustration}.

Our main contributions are summarized as follows:
\begin{itemize}
    \item We present \method, a training-free and scalable framework for
    temporally consistent and robust human mesh recovery from in-the-wild videos.
    \item We achieve identity-consistent 4D mesh trajectories across complex
    and dynamic scenes using temporally aligned masklets as prompts.
    \item We propose an occlusion-aware refinement mechanism that improves
    reconstruction quality under occlusions.
\end{itemize}

% Figure 1 (comparisons between SAM~3D~Body & Ours)

\section{Related Work}

\noindent\textbf{Human Mesh Recovery.}
% image HMR -> video HMR(解决image level的时空连续性) [同transformer/RNN时序extractor，或者用运动loss，或者其他模型] -> 但是都需要训练，我们提出一种novel的方式 用training-free的方式达到了稳定HMR in wild video的效果
Image-based human mesh recovery (HMR) predicts 3D human body meshes directly from a RGB image.
Existing approaches can be broadly categorized into regression-based methods and token-based methods.
The former directly regresses parameters of 3D human models, such as SMPL-X~\cite{SMPLX}, from image features (\textit{e.g.}, HMR~\cite{hmr}, SPIN~\cite{SPIN}, and PromptHMR~\cite{prompthmr}) while the latter represent joints or mesh vertices as learnable tokens and employ transformer reasoning to model their relationships, enabling more flexible and expressive structured prediction (\textit{e.g.} TokenHMR~\cite{tokenhmr} and MEGA~\cite{mega}).

Despite the strong performance of image-based HMR methods, they do not naturally adapt to video settings, where challenges such as temporal consistency and occlusions become critical.
To enforce temporal smoothness, feature-level temporal modeling approaches, such as VIBE~\cite{vibe} and TRAM~\cite{tram}, build upon image-based feature encoders and introduce temporal modules such as GRU and transformers to aggregate frame-wise representations and learn coherent motion cues over time.
4DHumans~\cite{human4d}, on the other hand, jointly performs mesh reconstruction and identity-consistent tracking, enabling stable human modeling under occlusions and rapid pose variation.
% TODO: 在源头预测;而不是结果
In addition, these approaches are fundamentally optimization-based, requiring large amounts of manually annotated video data and carefully designed loss functions, which restrict their generalizability and scalability.
In contrast, our method is entirely training-free, enforcing temporal coherence
from the source by directly leveraging the pixel-level continuity of humans
in video, rather than relying on feature- or pose-space temporal modelling
where such continuity may already be lost.

% {\color{blue}add some introduction of method}, offering strong flexibility and generalization.

\noindent\textbf{Video Object Segmentation (VOS)} focuses on tracking and segmenting a target object across video frames, typically based on a few pixel/box annotations~\cite{gao2023deep} or text prompts~\cite{zhou2022survey}. Prior VOS methods generally follow a memory-based paradigm~\cite{stm,xmem,cutie}, where historical predictions are leveraged to maintain temporal consistency. However, these approaches still struggle to generalize to in-the-wild scenarios due to limited model capacity and training diversity.

Benefiting from billion-scale training data and strong transformer architectures, SAM~\cite{sam} demonstrates high segmentation accuracy, strong generalisation to in-the-wild images, and rich prompting modalities.
By introducing a memory mechanism, SAM~2~\cite{sam2} extends these strengths to the VOS setting.
More recently, SAM~3~\cite{sam3} incorporates text prompts and an independent object perceiver, enabling more user-friendly interactions and stronger robustness to common challenges such as disappearance and reappearance in complex videos.

Despite strong performance on visible regions, SAM-based VOS methods cannot recover occluded body parts, resulting in incomplete visual cues for downstream HMR and causing hallucinated geometry during occlusion.
To address this limitation, we introduce an occlusion-aware refinement module that reconstructs hidden regions and provides full-body references for robust and consistent HMR in videos.

\begin{figure*}[ht]
  \centering
  \includegraphics[width=1\linewidth]{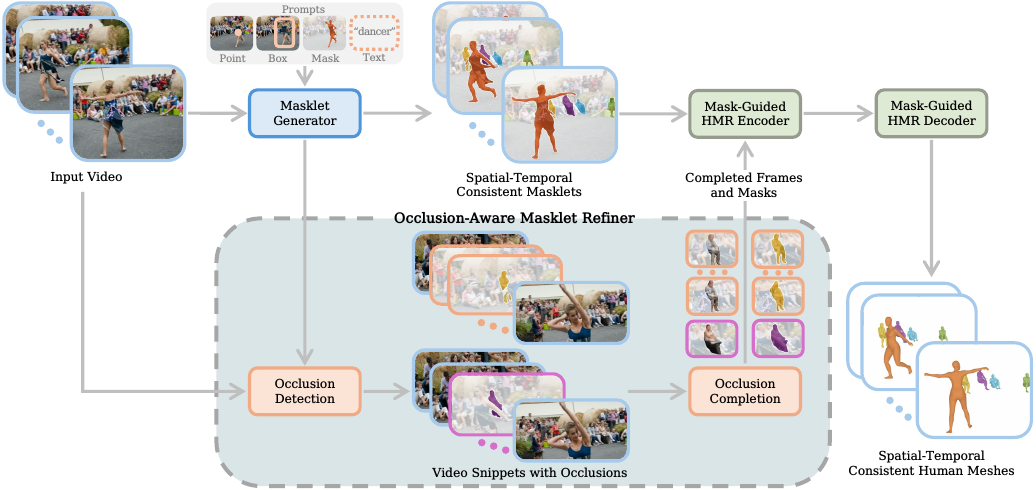}
  \caption{Overall framework of the proposed \method. Given an input video with human prompts, \method operates on three main modules in a training-free manner. The \textcolor{mask}{Masklet Generator} derives identity-consistent temporal masklets from the video to provide spatio-temporal tracking cues. The \textcolor{occlusion}{Occlusion-Aware Masklet Refiner} enriches these masklets by recovering invisible body regions and stabilizing temporal alignment. Finally, the \textcolor{hmr}{Mask-Guided HMR module} uses refined masklets as spatial prompts to predict accurate and temporally coherent human meshes across the entire sequence. }
  \label{fig:framework}
\end{figure*}

\section{Preliminary}
\noindent \textbf{SAM~3}~\cite{sam3} is a promptable object segmentation model supporting both images and videos. 
Given a video 
$\mathcal{V} = \{I_t\}_{t=1}^T$, $I_t \in \mathbb{R}^{H \times W \times 3}$,
and user-defined prompts $\mathcal{P}$, SAM~3 predicts a mask sequence 
$\mathcal{M} = \{M_t\}_{t=1}^T$, $M_t \in \{0,1\}^{H \times W}$.

For the $t$-th frame, the mask is obtained by combining two complementary modules: 
\texttt{propagate} and \texttt{detect}.
\begin{equation}
\label{eq:sam3}
\begin{aligned}
\hat{M_t} &= \text{propagate}(M_{t-1}), \\
O_t &= \text{detect}(I_t, \mathcal{P}), \\
M_t &= \text{match\_and\_update}(\hat{M_t}, O_t).
\end{aligned}
\end{equation}
The \texttt{propagate} module leverages spatial-temporal correspondences between historical predictions 
(stored in memory) and the current frame, enabling reliable transfer of mask labels across time and preserving 
target continuity throughout the video. In contrast, the \texttt{detect} module focuses on semantic associations 
between the prompt and objects in the current frame, which is particularly effective for challenging cases such 
as disappearance and reappearance in complex scenes. By integrating the outputs of the two complementary modules through \texttt{match\_and\_update}, 
SAM~3 achieves substantially higher accuracy than SAM~2 and other VOS methods on complex videos. 

\noindent \textbf{SAM~3D~Body}~\cite{sam3dbody} is a promptable model supporting Human Mesh Recovery (HMR) from in-the-wild images. 
It accepts prompts at both the encoder and decoder stages. Given an input image $I \in \mathbb{R}^{H \times W \times 3}$, 
SAM~3D~Body performs feature encoding as:
\begin{equation}
\label{eq:encoder}
F = \text{ImgEncoder}(I, \mathcal{P}_{\text{enc}}),
\end{equation}
where $\mathcal{P}_{\text{enc}}$ denotes optional encoder prompts (e.g., 2D keypoints or segmentation masks) 
that help the model focus on the target human. With encoded features, the decoder predicts full-body tokens as:
\begin{equation}
\label{eq:decoder}
O = \text{Decoder}(F, \mathcal{P}_{\text{dec}}),
\end{equation}
where $\mathcal{P}_{\text{dec}}$ includes optional prompts such as keypoint, camera, or MHR tokens. Then the first output token $O_0$ is passed through an MLP to obtain MHR parameters:
\begin{equation}
\label{eq:mhr}
\theta = \{P, S, C, S_k\} = \text{MLP}(O_0),
\end{equation}
where $P$, $S$, $C$, and $S_k$ denote pose, shape, camera pose, and skeleton parameters.

During inference, body and hand are optimised separately due to different optimisation strategies and later 
fused. Unless otherwise specified, $\theta$ indicates the final full-body mesh parameters containing both body and hand information.

\section{Methodology}
% Framework

\subsection{Overview}
The framework of \method is illustrated in Fig.~\ref{fig:framework}. 
Given an input video $\mathcal{V} = \{I_t\}_{t=1}^T$ and prompts 
$\mathcal{P} = \{\mathcal{P}^{h_i}\}_{i=1}^N$ indicating $N$ target humans, 
\method estimates temporally consistent human mesh parameters 
$\Theta = \{\theta_t^{h_i}\}_{t=1,i=1}^{T,N}$ for all selected persons.

\method consists of three key components. A \textit{Masklet Generator} produces 
identity-consistent masklets as temporal tracking cues. An \textit{Occlusion-Aware Masklet Refiner} enhances these masklets by recovering missing regions when occlusion occurs. A \textit{Mask-Guided HMR module} then predicts per-frame mesh parameters $\theta_t$. Since each mesh is aligned with its corresponding mask over time, the temporal continuity in masklets is naturally propagated to the reconstructed human meshes.

\subsection{Masklet Generator}

For each target human $h_i$ specified by prompts $\mathcal{P}$, 
we apply SAM~3~\cite{sam3} over the video $\mathcal{V}$ to obtain spatio-temporally 
aligned masklets $\mathcal{M} = \{M_t^{h_i}\}$. Following the hybrid 
propagation-detection formulation in Eq.~\ref{eq:sam3}, identity consistency 
is maintained across frames. Note that other video segmentation models capable 
of producing temporally aligned instance masks can also be adopted here. 

\subsection{Occlusion-Aware Masklet Refiner}

In in-the-wild videos, humans frequently undergo severe occlusions, where even 
state-of-the-art segmentation methods like SAM~3 can only capture visible body 
regions. Such incomplete masklets provide insufficient visual evidence for 
human mesh estimation and may lead to hallucinated predictions with unrealistic pose 
and shape. To resolve this issue, we introduce an occlusion-aware refinement 
module to recover missing regions and offer complete visual references for the 
subsequent HMR stage.

We first obtain mask completion results by feeding video frames $\mathcal{V}$ 
and masklets $\mathcal{M}$ into a mask completion model (Diffusion-VAS~\cite{diffusion_vas}), 
producing completed masklets $\tilde{\mathcal{M}} = \{\tilde{M}_t^{h_i}\}$. 
For each human $h_i$ at frame $t$, occlusions are detected when the completed 
mask area becomes larger while the overlap remains low:
\begin{equation}
\label{eq:occ_detect}
\text{occ}(t,h_i) = 
\mathbbm{1}\!\left(
\begin{aligned}
&|\tilde{M}_t^{h_i}| > |M_t^{h_i}| \\
&\wedge\; \text{IoU}(\tilde{M}_t^{h_i}, M_t^{h_i}) < 0.7
\end{aligned}
\right)
\end{equation}
To obtain accurate visual evidence for the occluded regions, the detected frames 
are temporally grouped and re-fed to Diffusion-VAS to recover missing pixels 
(see the yellow module in Fig.~\ref{fig:framework}). We then update the corresponding frames and 
masks as:
\begin{equation}
\label{eq:occ_replace}
I_t^{h_i} \leftarrow \tilde{I}_t^{h_i},\quad 
M_t^{h_i} \leftarrow \tilde{M}_t^{h_i},
\end{equation}
yielding refined video $\tilde{\mathcal{V}}$ and refined masklets 
$\tilde{\mathcal{M}}$ with full-body visual cues and improved temporal 
stability, which provide more reliable supervision for the subsequent HMR 
module and enable more accurate and consistent meshes across challenging frames.

\begin{figure*}[!t]

  \centering
  \includegraphics[width=1\linewidth]{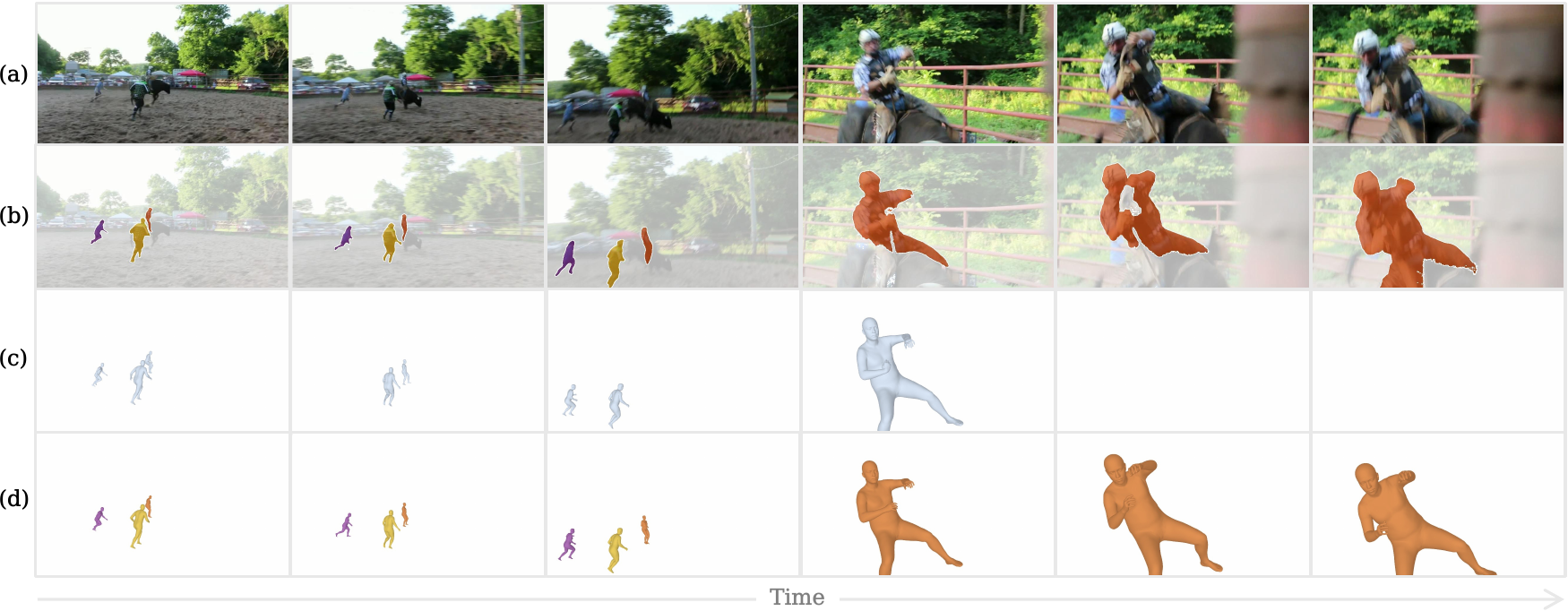}
  \caption{Visualised comparisons between the vanilla image-to-video extension of SAM~3D-Body and our \method. (a) Input video frames. (b) Identity-consistent human masks. (c) Vanilla per-frame HMR results using SAM~3D-Body with automatic human detection, where missed detections lead to missing meshes. (d) Our \method maintains temporally continuous and identity-preserving mesh trajectories throughout the video by leveraging spatial-temporal masklet guidance.}
  \label{fig:vanilla_comparison}
\end{figure*}

\begin{figure*}[!t]
  \centering
  \includegraphics[width=1\linewidth]{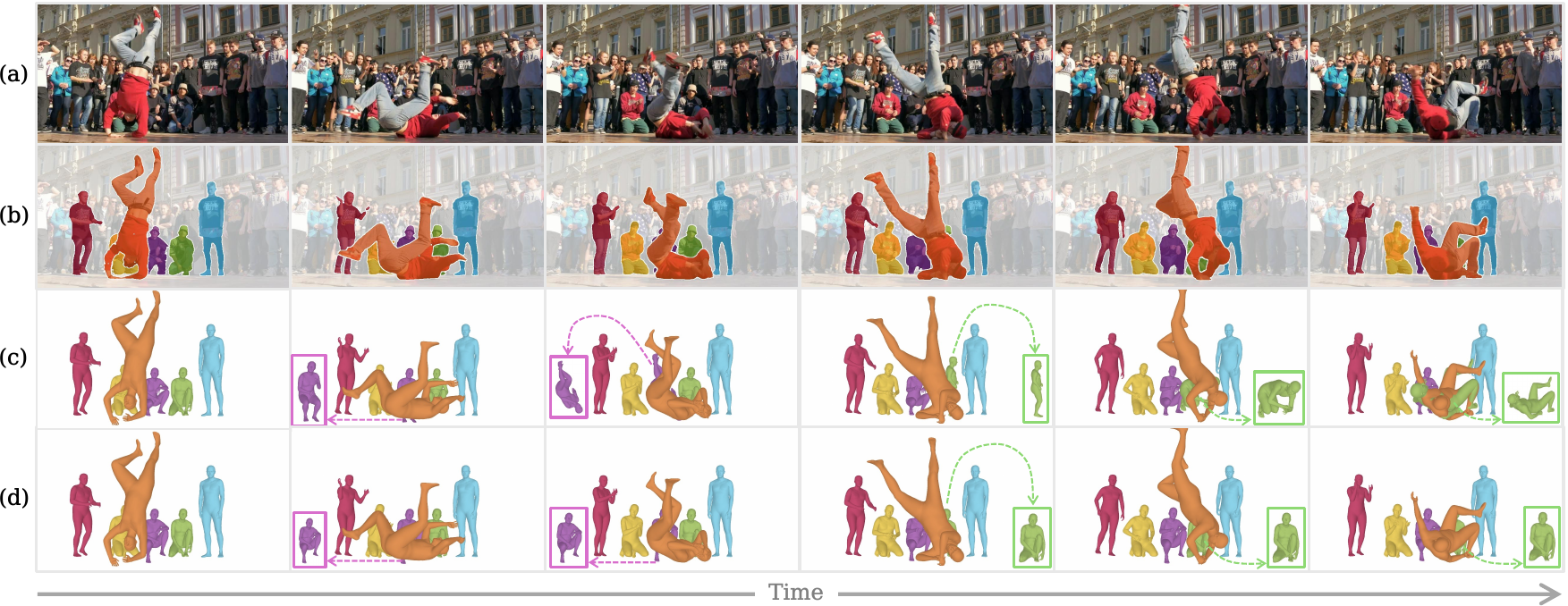}
  \caption{Visualised comparisons between \method w/o and w/ Occlusion-Aware Masklet Refiner. (a) Input video frames; (b) Temporally consistent human masks, where each person is highlighted with a unique and consistent color across frames; (c) \method without Occlusion-Aware Masklet Refiner; (d) \method with Occlusion-Aware Masklet Refiner. Across the 2nd–6th columns, \method produces more robust reconstructions under occlusion (e.g., the blue-rendered person in the 2nd column, the purple-rendered people in the 3rd/4th column, and the green-rendered people in the 5th and 6th columns). Since these subjects are heavily occluded, their meshes without occlusion are shown at the bottom-left/bottom-right for clearer observation.}
  \label{fig:occ_comparison}
\end{figure*}

\begin{algorithm}[t]
\small
  \caption{Training-Free Mask-Guided Video HMR}
  \label{alg:method}
  \begin{algorithmic}
    \STATE \textbf{Input:} Video $\mathcal{V} = \{I_t\}_{t=1}^T$, prompts $\mathcal{P} = \{\mathcal{P}^{h_i}\}_{i=1}^N$
    \STATE \textbf{Output:} Mesh parameters $\Theta = \{\theta_t^{h_i}\}_{t=1,i=1}^{T,N}$
    \STATE

    \STATE \textbf{// 1. Masklet Generation (SAM~3)}
    \STATE Obtain initial masklets $\mathcal{M} = \{M_t^{h_i}\}$ using SAM~3 via Eq.~\ref{eq:sam3}.

    \STATE \textbf{// 2. Occlusion-Aware Refinement (Diffusion-VAS)}
    \STATE Complete masks using Diffusion-VAS to obtain $\tilde{\mathcal{M}} = \{\tilde{M}_t^{h_i}\}$.
    \FOR{each frame $t$ and human $h_i$}
      \IF{occlusion $\text{occ}(t,h_i)$ detected by Eq.~\ref{eq:occ_detect}}
        \STATE Recover missing pixels and update video and masks via Eq.~\ref{eq:occ_replace}.
      \ENDIF
    \ENDFOR
    \STATE Obtain refined video $\tilde{\mathcal{V}}$ and refined masklets $\tilde{\mathcal{M}}$.

    \STATE

    \STATE \textbf{// 3. Parallel Mask-Guided HMR (SAM~3D~Body)}
    \STATE Construct frame batches and pad missing humans to a fixed batch shape.
    \FOR{each frame batch}
      \STATE Use $\tilde{M}_t^{h_i}$ as encoder prompts $\mathcal{P}_{\text{enc}}$.
      \STATE Compute $\theta_t^{h_i}$ via Eq.~\ref{eq:encoder} and Eq.~\ref{eq:decoder} in a single forward pass.
    \ENDFOR

    \STATE

    \STATE \textbf{// 4. Temporal Smoothing in MHR Space}
    \FOR{each human $h_i$}
      \STATE Fix scale and shape to those from the first visible frame of $h_i$.
      \STATE Apply Kalman smoothing to the per-frame pose and hand parameters of $h_i$.
    \ENDFOR

    \STATE

    \STATE \textbf{return} $\Theta$
  \end{algorithmic}
\end{algorithm}

\subsection{Training-Free Mask-Guided HMR}

With refined video $\tilde{\mathcal{V}}$ and refined masklets 
$\tilde{\mathcal{M}}$, we perform training-free HMR. For each target human
$h_i$, the corresponding mask $\tilde{M}_t^{h_i}$ is used as the encoder
prompt $\mathcal{P}_{\text{enc}}$ in Eq.~\ref{eq:encoder} and
Eq.~\ref{eq:decoder}, enabling the model to focus on the correct identity.
This produces per-frame mesh parameters 
$\Theta = \{\theta_t^{h_i}\}_{t=1,i=1}^{T,N}$ consistent with the refined
masklets over time. Importantly, the entire pipeline operates in a
\textbf{training-free} manner without any task-specific finetuning.

We further improve the efficiency of the SAM~3D~Body HMR stage. The original
pipeline performs per-frame sequential inference, and the number of visible
humans may vary across frames, making naive batching infeasible. We introduce
a simple padding mechanism to unify the batch shape so that all humans within
the same frame batch can be processed jointly in a single forward pass. This
parallelisation eliminates redundant per-human inference and yields a
substantial speed-up while preserving the temporal alignment of the predicted
meshes.

To further enhance motion stability, we apply lightweight test-time temporal
smoothing to the Momentum Human Rig pose and hand parameters, reducing jitter
and promoting smooth transitions. In addition, for each target human, the
scale and shape parameters from the first visible frame are reused across the
entire sequence to maintain consistent body proportions and avoid identity
drift. These operations require no learning and introduce negligible
computational overhead, keeping the entire framework fully \textbf{training-free}.

The full procedure of \method is summarised in
Algorithm~\ref{alg:method}, which complements the structural illustration in
Fig.~\ref{fig:framework} by explicitly detailing the execution flow,
including identity-consistent masklet generation, occlusion-aware refinement,
and our parallel mask-guided HMR strategy.

\section{Experiments}

This section presents qualitative results to demonstrate the effectiveness of our training-free framework for video HMR. Our implementation combines SAM~3~\cite{sam3} as the Masklet Generator, Diffusion-VAS~\cite{diffusion_vas} for occlusion detection and refinement, and SAM~3D~Body~\cite{sam3dbody} for Mask-Guided HMR. The detection and refinement stages operate at a spatial resolution of $512 \times 1024$, while lower resolutions can further improve efficiency with a moderate loss in visual fidelity.

The pipeline supports efficient deployment on a single GPU. Without the occlusion refiner, our parallel multi-frame inference achieves substantial speed improvements over sequential per-frame HMR. For instance, on an NVIDIA A100-SXM4-80GB (96GB system memory), processing a $480 \times 854$ video (90 frames, 5 persons) runs approximately $2\times$ faster with a parallel
batch size of 32. When the refiner is enabled, both memory usage and runtime increase depending on the duration of the occlusion and the number of persons being refined.

\subsection{Comparison with Vanilla Image-to-Video Extension}
\label{sec:vanilla_compare}

We evaluate the vanilla image-to-video extension of SAM 3D Body, where mesh
prediction is performed independently on each frame without any temporal enforcement. As shown in Fig.~\ref{fig:vanilla_comparison}, such a per-frame
strategy cannot ensure continuous mesh trajectories for the same person across
the video. When the target becomes small, suffers motion blur, or is heavily
occluded, the detector may fail to localise the human, leading to missing
meshes.

In contrast, \method leverages temporally aligned masklets to provide
consistent target localisation throughout the sequence. These masklets are used
as element-wise prompts for HMR, effectively transferring pixel-level
continuity to 4D human meshes. Owing to the one-to-one correspondence between
mask regions and reconstructed meshes, identity association remains stable,
even when visibility temporarily degrades.

This comparison highlights that enforcing spatial-temporal continuity at the
pixel level is essential for reliable and stable video HMR. By preserving
consistent localisation cues, \method maintains identity association and smooth
mesh evolution across frames, enabling coherent 4D reconstruction throughout
the video.

\subsection{Effectiveness of Occlusion-Aware Refinement}
\label{sec:occ_effect}

Occlusions occur frequently in real-world videos, where major body areas are
temporarily hidden by objects or other people. In such cases, per-frame HMR
relies on incomplete visual evidence and easily hallucinates implausible body
structures. To demonstrate the benefit of our refinement, we present visual
comparisons under challenging occlusion scenarios in
Fig.~\ref{fig:occ_comparison}.

When only a small portion of the body is occluded (e.g., first column of
Fig.~\ref{fig:occ_comparison}), the vanilla SAM~3D~Body can still produce
reasonable predictions. However, once most of the body becomes invisible, the
baseline depends solely on the limited visible pixels and generates distorted
and unstable meshes. In contrast, our occlusion-aware refiner restores missing
human regions before HMR inference, enabling \method to preserve plausible pose
and consistent body structure throughout the occluded frames.

These results highlight that recovering occluded body evidence is essential to
mitigate hallucinated predictions and achieve reliable 4D human reconstruction
in challenging in-the-wild videos.

\FloatBarrier
\section{Conclusion}

We presented \method, a training-free framework for temporally consistent
Human Mesh Recovery (HMR) from videos. By leveraging identity-consistent
masklets and an occlusion-aware refinement module, our approach effectively
transfers pixel-level continuity into coherent 4D human mesh reconstruction.
Without requiring any additional training or architectural modification to
SAM-3D-Body, \method improves temporal stability and robustness in challenging
in-the-wild scenarios. Our parallel multi-frame inference strategy further
enables efficient and scalable deployment in practical applications.

\nocite{langley00}

\bibliography{example_paper}

@String(PAMI = {IEEE Trans. Pattern Anal. Mach. Intell.})

@String(CVPR= {IEEE Conf. Comput. Vis. Pattern Recog.})

@String(ICCV= {Int. Conf. Comput. Vis.})

@String(ECCV= {Eur. Conf. Comput. Vis.})

@String(ICLR = {Int. Conf. Learn. Represent.})

@String(PAMI  = {IEEE TPAMI})

@String(CVPR  = {CVPR})

@String(ICCV  = {ICCV})

@String(ECCV  = {ECCV})

@String(ICLR  = {ICLR})

@misc{sam3,
      title={SAM 3: Segment Anything with Concepts},
      author={Nicolas Carion and Laura Gustafson and Yuan-Ting Hu and Shoubhik Debnath and Ronghang Hu and Didac Suris and Chaitanya Ryali and Kalyan Vasudev Alwala and Haitham Khedr and Andrew Huang and Jie Lei and Tengyu Ma and Baishan Guo and Arpit Kalla and Markus Marks and Joseph Greer and Meng Wang and Peize Sun and Roman Rädle and Triantafyllos Afouras and Effrosyni Mavroudi and Katherine Xu and Tsung-Han Wu and Yu Zhou and Liliane Momeni and Rishi Hazra and Shuangrui Ding and Sagar Vaze and Francois Porcher and Feng Li and Siyuan Li and Aishwarya Kamath and Ho Kei Cheng and Piotr Dollár and Nikhila Ravi and Kate Saenko and Pengchuan Zhang and Christoph Feichtenhofer},
      year={2025},
      eprint={2511.16719},
      archivePrefix={arXiv},
      primaryClass={cs.CV},
      url={https://arxiv.org/abs/2511.16719},
}

@article{sam3dbody,
  title={SAM 3D Body: Robust Full-Body Human Mesh Recovery},
  author={Yang, Xitong and Kukreja, Devansh and Pinkus, Don and Sagar, Anushka and Fan, Taosha and Park, Jinhyung and Shin, Soyong and Cao, Jinkun and Liu, Jiawei and Ugrinovic, Nicolas and Feiszli, Matt and Malik, Jitendra and Dollar, Piotr and Kitani, Kris},
  journal={arXiv preprint; identifier to be added},
  year={2025}
}

@inproceedings{SMPLX,
    title = {Expressive Body Capture: 3D Hands, Face, and Body from a Single Image},
    author = {Pavlakos, Georgios and Choutas, Vasileios and Ghorbani, Nima and Bolkart, Timo and Osman, Ahmed A. A. and Tzionas, Dimitrios and Black, Michael J.},
    booktitle = {CVPR},
    year = {2019}
}

@article{mhr,
  title={MHR: Momentum Human Rig},
  author={Ferguson, Aaron and Osman, Ahmed AA and Bescos, Berta and Stoll, Carsten and Twigg, Chris and Lassner, Christoph and Otte, David and Vignola, Eric and Bogo, Federica and Santesteban, Igor and others},
  journal={arXiv preprint arXiv:2511.15586},
  year={2025}
}

@inproceedings{diffusion_vas,
  title={Using Diffusion Priors for Video Amodal Segmentation},
  author={Chen, Kaihua and Ramanan, Deva and Khurana, Tarasha},
  booktitle={CVPR},
  pages={22890--22900},
  year={2025}
}

@inproceedings{hmr,
  title={End-to-end recovery of human shape and pose},
  author={Kanazawa, Angjoo and Black, Michael J and Jacobs, David W and Malik, Jitendra},
  booktitle={CVPR},
  pages={7122--7131},
  year={2018}
}

@inproceedings{SPIN,
  title={Learning to estimate 3D human pose and shape from a single color image},
  author={Pavlakos, Georgios and Zhu, Luyang and Zhou, Xiaowei and Daniilidis, Kostas},
  booktitle={CVPR},
  pages={459--468},
  year={2018}
}

@inproceedings{prompthmr,
  title={PromptHMR: Promptable Human Mesh Recovery},
  author={Wang, Yufu and Sun, Yu and Patel, Priyanka and Daniilidis, Kostas and Black, Michael J and Kocabas, Muhammed},
  booktitle={CVPR},
  pages={1148--1159},
  year={2025}
}

@inproceedings{tokenhmr,
  title={Tokenhmr: Advancing human mesh recovery with a tokenized pose representation},
  author={Dwivedi, Sai Kumar and Sun, Yu and Patel, Priyanka and Feng, Yao and Black, Michael J},
  booktitle=CVPR,
  pages={1323--1333},
  year={2024}
}

@inproceedings{vibe,
  title={Vibe: Video inference for human body pose and shape estimation},
  author={Kocabas, Muhammed and Athanasiou, Nikos and Black, Michael J},
  booktitle={CVPR},
  pages={5253--5263},
  year={2020}
}

@inproceedings{tram,
  title={TRAM: Global Trajectory and Motion of 3D Humans from in-the-wild Videos},
  author={Wang, Yufu and Wang, Ziyun and Liu, Lingjie and Daniilidis, Kostas},
  booktitle={ECCV},
  pages={467--487},
  year={2024},
  organization={Springer}
}

@inproceedings{mega,
  title={MEGA: Masked Generative Autoencoder for Human Mesh Recovery},
  author={Fiche, Gu{\'e}nol{\'e} and Leglaive, Simon and Alameda-Pineda, Xavier and Moreno-Noguer, Francesc},
  booktitle=CVPR,
  pages={5366--5378},
  year={2025}
}

@inproceedings{human4d,
  title={Humans in 4d: Reconstructing and tracking humans with transformers},
  author={Goel, Shubham and Pavlakos, Georgios and Rajasegaran, Jathushan and Kanazawa, Angjoo and Malik, Jitendra},
  booktitle={ICCV},
  pages={14783--14794},
  year={2023}
}

@article{gao2023deep,
  title={Deep learning for video object segmentation: a review},
  author={Gao, Mingqi and Zheng, Feng and Yu, James JQ and Shan, Caifeng and Ding, Guiguang and Han, Jungong},
  journal={Artificial Intelligence Review},
  volume={56},
  number={1},
  pages={457--531},
  year={2023},
  publisher={Springer}
}

@article{zhou2022survey,
  title={A survey on deep learning technique for video segmentation},
  author={Zhou, Tianfei and Porikli, Fatih and Crandall, David J and Van Gool, Luc and Wang, Wenguan},
  journal=PAMI,
  volume={45},
  number={6},
  pages={7099--7122},
  year={2022},
  publisher={IEEE}
}

@inproceedings{stm,
  title={Video object segmentation using space-time memory networks},
  author={Oh, Seoung Wug and Lee, Joon-Young and Xu, Ning and Kim, Seon Joo},
  booktitle=ICCV,
  pages={9226--9235},
  year={2019}
}

@inproceedings{xmem,
  title={Xmem: Long-term video object segmentation with an atkinson-shiffrin memory model},
  author={Cheng, Ho Kei and Schwing, Alexander G},
  booktitle=ECCV,
  pages={640--658},
  year={2022},
  organization={Springer}
}

@inproceedings{cutie,
  title={Putting the object back into video object segmentation},
  author={Cheng, Ho Kei and Oh, Seoung Wug and Price, Brian and Lee, Joon-Young and Schwing, Alexander},
  booktitle=CVPR,
  pages={3151--3161},
  year={2024}
}

@inproceedings{sam,
  title={Segment anything},
  author={Kirillov, Alexander and Mintun, Eric and Ravi, Nikhila and Mao, Hanzi and Rolland, Chloe and Gustafson, Laura and Xiao, Tete and Whitehead, Spencer and Berg, Alexander C and Lo, Wan-Yen and others},
  booktitle=ICCV,
  pages={4015--4026},
  year={2023}
}

@inproceedings{sam2,
  title={Sam 2: Segment anything in images and videos},
  author={Ravi, Nikhila and Gabeur, Valentin and Hu, Yuan-Ting and Hu, Ronghang and Ryali, Chaitanya and Ma, Tengyu and Khedr, Haitham and R{\"a}dle, Roman and Rolland, Chloe and Gustafson, Laura and others},
  booktitle=ICLR,
  year={2024}
}
\bibliographystyle{icml2026}

%%%%%%%%%%%%%%%%%%%%%%%%%%%%%%%%%%%%%%%%%%%%%%%%%%%%%%%%%%%%%%%%%%%%%%%%%%%%%%%
%%%%%%%%%%%%%%%%%%%%%%%%%%%%%%%%%%%%%%%%%%%%%%%%%%%%%%%%%%%%%%%%%%%%%%%%%%%%%%%
% APPENDIX
%%%%%%%%%%%%%%%%%%%%%%%%%%%%%%%%%%%%%%%%%%%%%%%%%%%%%%%%%%%%%%%%%%%%%%%%%%%%%%%
%%%%%%%%%%%%%%%%%%%%%%%%%%%%%%%%%%%%%%%%%%%%%%%%%%%%%%%%%%%%%%%%%%%%%%%%%%%%%%%
% \newpage
% \appendix
% \onecolumn
% \section{You \emph{can} have an appendix here.}

% You can have as much text here as you want. The main body must be at most $8$
% pages long. For the final version, one more page can be added. If you want, you
% can use an appendix like this one.

% The $\mathtt{\backslash onecolumn}$ command above can be kept in place if you
% prefer a one-column appendix, or can be removed if you prefer a two-column
% appendix.  Apart from this possible change, the style (font size, spacing,
% margins, page numbering, etc.) should be kept the same as the main body.
%%%%%%%%%%%%%%%%%%%%%%%%%%%%%%%%%%%%%%%%%%%%%%%%%%%%%%%%%%%%%%%%%%%%%%%%%%%%%%%
%%%%%%%%%%%%%%%%%%%%%%%%%%%%%%%%%%%%%%%%%%%%%%%%%%%%%%%%%%%%%%%%%%%%%%%%%%%%%%%

\end{document}